# CovidMis20: COVID-19 Misinformation Detection System on Twitter Tweets using Deep Learning Models

Aos Mulahuwaish[1], Manish Osti[1], Kevin Gyorick[1], Majdi Maabreh[2], Ajay Gupta[3], and Basheer Qolomany[4]

[1] Department of Computer Science and Information Systems, Saginaw Valley State University, University Center, USA
amulahuw@svsu.edu, mrosti@svsu.edu, kpgyoric@svsu.edu
[2]Department of Information Technology, Faculty of Prince Al-Hussein Bin Abdallah II for Information Technology, The Hashemite University, Zarqa, Jordan
majdi@hu.edu.jo
[3]Department of Computer Science, Western Michigan University, Kalamazoo, USA
ajay.gupta@wmich.edu
[4] Cyber Systems Department, University of Nebraska at Kearney, Kearney, USA
qolomanyb@unk.edu

**Abstract.** Online news and information sources are convenient and accessible ways to learn about current issues. For instance, more than 300 million people engage with posts on Twitter globally, which provides the possibility to disseminate misleading information. There are numerous cases where violent crimes have been committed due to fake news. This research presents the CovidMis20 dataset (COVID-19 Misinformation 2020 dataset), which consists of 1,375,592 tweets collected from February to July 2020. CovidMis20 can be automatically updated to fetch the latest news and is publicly available at: https://github.com/everythingguy/CovidMis20.

This research was conducted using Bi-LSTM deep learning and an ensemble CNN+Bi-GRU for fake news detection. The results showed that, with testing accuracy of 92.23% and 90.56%, respectively, the ensemble CNN+Bi-GRU model consistently provided higher accuracy than the Bi-LSTM model.

**Keywords:** COVID-19, Misinformation, Dataset, Benchmark, Fake News, Twitter, Deep Learning.

## 1 Introduction

Fake news, misinformation and disinformation have been in the media for a long time, with instances cited as far back as 1835 [1]. However, instances of fake news have grown detrimental with the growth of the internet, especially through social media platforms like Twitter, Facebook, etc. Concerns about fake news grew exponentially during the COVID-19 pandemic and the 2020 US Presidential Election. When the pandemic started in 2019, worry increased, and people went to the internet for answers. Millions of people died from COVID-19, and the global pandemic had lasting economic effects.

In addition to COVID-19, there was a US Presidential election in 2020. These two events led to increased fake news led by political biases.

Cases of fake news were especially prevalent on social media platforms, one such social media platform being Twitter. Twitter is a platform where people can share information and images in "Tweets." However, since virtually anybody can post anything, misinformation can also be posted and spread quickly. Therefore, we need to have a way to flag content that could be untrustworthy.

To classify the CovidMis20 dataset, we chose to use Ensemble Convolutional Neural Network (CNN) with the Bi-Directional Gated Recurrent Unit (Bi-GRU) model because Bi-GRU is a less complex architecture than Bidirectional Long Short-Term Memory (Bi-LSTM) [2], which is a good model for text classification purposes in general. CNN is better suited to process the higher dimensional matrices, whereas Bi-GRU is better at temporal sequence data. So, using CNN and Bi-GRU as an ensemble model, we can utilize the power of calculation of higher dimensional matrices and temporal sequence data with less complexity than Bi-LSTM. An alternative is to use Bi-LSTM with Bi-GRU, but we will later see that the ensemble of CNN to capture the high dimensional matrices with Bi-GRU capturing temporal features performed better.

We could also try to use classical machine learning techniques like Support Vector Machine (SVM), k-nearest neighbors (k-NN), Decision Trees, etc. However, we believe deep learning has far more capability in generalizing the various datasets and can be improved with further improvements in the model's architectures.

The main contributions of this work are as follows:
- We built a labeled dataset, CovidMis20, which contains around 1,375,592 tweets; this dataset helps researchers differentiate between fake and real news related to COVID-19 on Twitter; also, CovidMis20 can be automatically updated to fetch the latest news. We collected the Twitter dataset and used Media Bias Fact Check (MBFC) [16], which is a fact-checking page that relies strictly on the International Fact-Checking Network (IFCN) signatories. MBFC portal has one of the most comprehensive databases of fact-checking media sources. Initially, a dataset of URLs identified as fake or real was created using a web crawler tool to identify URLs with COVID-19 information. Based on the Media Bias Fact Check (MBFC) database, these URLs were classified as fake and real.
- We developed and evaluated two deep learning models: ensemble CNN+Bi-GRU and Bi-LSTM for fake and real tweets text prediction with a testing accuracy of 92.23% and 90.56%, respectively.

This paper is organized as follows. Section 2 discusses related works. Section 3 presents the methodology and implementation. Section 4 presents results and discussion. Finally, Section 5 provides our conclusions and future work.

## 2  Related Work

Recently, a significant amount of work has been done in the area of fake news detection. This section reviews recent related works on the different aspects of fake news detection.

Hassan et al. [3] proposed a machine learning approach using supervised machine learning classification models like Support Vector Machine (SVM), Logistic Regression, and Naive Bayes to detect fake online reviews. They have used 1600 examples of the gold standard hotel review dataset, 800 reviews are deceptive, and the other 800 are truthful; as a result, they found that the SVM has given the best accuracy (88.75%) over the remaining models.

Reddy et al. [4] introduced a hybrid detection system for fake data that employs the Multinomial Voting Algorithm based on Naïve Bayes, Random Forest, Decision Tree, Support Vector Machine (SVM), k-Nearest Neighbours (k-NN) models. Their experimental data was collected from the Kaggle datasets (the authors did not mention anything related to the type of the datasets and their sizes or features). They had an accuracy score of 92.58% using above mentioned classical machine learning techniques.

Patwa et al. [5] created a COVID-19 social media and articles dataset for real and fake labels. Fake claims are collected from different fact-checking websites (like Politifact2, NewsChecker 3, Boomlive 4, etc.), and the real labels are collected from Twitter. The dataset vocabulary size is about 37,505. They used classical machine learning techniques like Decision Tree, Logistic Regression, Support Vector Machine (SVM), and Gradient Boost, the SVM model, which performs best with 93.32% accuracy, which is also a very good precision, recall, and accuracy overall. However, SVM gets more complex to work with when increasing the dataset size.

Al Asaad et al. [6] introduced a model to verify the news credibility that used different machine learning techniques for text classification. They used two datasets (fake or real news) with 6335 news articles, 3164 were fake, and 3171 were true, and log data with 32000 titles, 15999 being clickbait and 16001 being non-clickbait. The model's efficiency has been tested on a dataset by using Multinomial Naïve Bayes (MN) and Linear Support Vector Machine classification (LSVC) algorithms; they applied them with Bag-of-Words (BoW 2), Bi-gram (bigram 3), and Term Frequency-Inverse Document Frequency: (TF-IDF). The LSVC classifier performs better with the TF-IDF model.

Yu et al. [7] proposed four hybrid deep learning models (CNN+GRU, CNN+Bi-RNN, CNN+Bi-LSTM, and CNN+Bi-GRU). They applied their models to a real dataset of patients' blood samples for the COVID-19 infection test from Hospital Israelita Albert Einstein in Sao Paulo, Brazil. They included 111 laboratory results from 5644 different patients. The results show that CNN+Bi-GRU performs the best, with an accuracy score of 94%.

Aslam et al. [8] introduced an ensemble Bi-LSTM+GRU dense, deep learning model to detect fake news where Bi-LSTM+GRU was implemented for the textual attribute. In contrast, a dense, deep learning model was used for the remaining attributes. The study used a LIAR dataset with 12.8 K humans labeled from POLITIFACT.COM. The proposed approach achieved an accuracy of 89.8%.

The "COVID-19 Fake News" dataset, which includes 21,379 real and fake news examples for the COVID-19 pandemic and associated vaccines, was used to train and test four deep neural networks for fake news detection. Convolutional Neural Network (CNN), Long Short Term Memory (LSTM), Bi-directional LSTM, and a combination of CNN and LSTM networks were developed and evaluated to automatically identify fake news content related to the COVID-19 posted on social media platforms. The evaluation's findings demonstrated that the CNN model performed better than the other deep neural networks, with an accuracy rate of 94.2% [9].

In summary, from the literature review, it is evident that social media platforms from the beginning of the pandemic were abuzz with news and information related to COVID-19. It was also clear that information – fake or real – caught the attention of millions of people active on social media, swayed their opinion, and influenced their behavior. To classify a tweet (that is, to analyze its content), many existing works analyze its text content using natural language processing (NLP) techniques and then use machine learning or deep learning algorithms for classifying the tweet's text. However, because the tweet's text is expressed using natural languages, it is hard to detect and extract what the tweet's author means due to the vagueness and the imprecision of the written text, which implies low accuracy.

To aid in those computational efforts, this paper describes the construction of the CovidMis20 (COVID Misinformation 2020), which could be used as a benchmark dataset; CovidMis20 contains confirmed true and fake news from Media Bias Fact Check (MBFC). The size of CovidMis20 is currently about 1,375,592 tweets; CovidMis20 can be automatically updated to fetch and incorporate the latest news. In contrast, most efforts in the literature have used relatively small datasets, and thus scalability remains a question. Obviously, most machine learning models make a good decision when the dataset size is relatively large. The results become more generalizable compared to smaller datasets. Hence, CovidMis20 should contribute to advancing the research on fake news detection. This paper also introduces a technique to build fake news detecting systems by focusing on creating a sequence classification model for fake (non-trustworthy) and real (trustworthy) tweets text (classifying text sequences based on the contextual information present).

## 3 Methodology and Implementation

### 3.1 Dataset Collection and Curation

We collected more than 1.5 billion coronavirus-related tweets from more than 40 million users from January 22, 2020, until May 15, 2022, leveraging the Twitter standard search application programming interface (API) and Tweepy Python library. A set of predefined search English keywords were used. These include {"corona," "coronavirus," "Coronavirus," "COVID-19", "stay at home," "lockdown," "Social Distancing," "Epidemic," "pandemic," and "outbreak"}, which are the some of the most widely used scientific and news media terms relating to the novel coronavirus. We extracted and stored the text and metadata of the tweets, such as timestamp, number of likes and retweets, hashtags, language, and user profile information, including user id, username, user location, number of followers, and number of friends.

Typically, it is challenging to label data as "fake" or "real" news since the perception varies from one individual to another. We can, however, focus on the facts that have been established medically or scientifically – and then decipher if these facts were incorrectly shared or twisted by individuals or websites while being posted on a platform.

Using a determined list of websites classified as trustworthy or not – obtained from Media Bias Fact Check (MBFC) - a web-crawler tool was deployed to collect an initial URL-based (MBFC dataset) dataset. The web crawler tool used in this work is a Java-based tool to crawl the internet. The tool was executed to search the internet and find and report COVID-19-related sites in an output file.

The keywords used for the search were – Sars-cov-2, COVID-19, Coronavirus, and virus – to only get sites related to the COVID-19 information. Then we looked for COVID articles at each URL (for example, https://achnews.org). If a page contained at least six instances of the key terms: Sars-cov-2, COVID-19, Coronavirus, and virus, it was considered a COVID article. The scraper went through each domain page, searching for these keywords. For the pages considered COVID articles, only the URL was collected. The website content needed to contain any of the keywords mentioned above thrice to ensure a closer hit. This ensured that the words were not present in the infomercial listed or the extra information section. There is an assumption made when we think only these keywords represent sites containing COVID-19 information. Of course, more sites have other keywords present talking about the virus. The URLs from the MBFC dataset were classified as fake and real as per the predetermined list from the MBFC.

For Twitter data analysis, applications were used to clean the data as the tweets that had MBFC were classified, URLs in them were extracted, and fields like tweet date, location, tweet text, and tweet URL were identified and extracted from the data for problem-solving.

Figure 1 shows the data collection process. The first phase of data collection is registering a Twitter application and obtaining a Twitter access key and token. The second phase is to import the Tweepy Python package and write the Python script for accessing Twitter API. The third phase is to connect to Twitter search API and retrieve tweets using some keywords related to COVID-19. The third phase reads and processes the data to extract information on tags, agents, and locations for network construction and analysis. The fourth phase filters Tweets containing URLs from a web crawler. The last phase is attaching MBFC labels, including trustworthy and untrustworthy ones.

### 3.2 Data Preprocessing

We used data mining and machine learning methods to preprocess, classify and analyze the data. Data or text mining, in general, helps identify relevant information in a large corpus, providing qualitative results, and machine learning for text mining or data mining is the process of reviewing the text to help with specific research questions. It is the initial cleaning part of data preprocessing. It helps identify the features and relationships of a given text. Once identified, these features can be used by machine learning methods to find patterns and trends across large data sets, resulting in more quantitative results. We used Natural Language Processing (NLP) which helps machines understand human language in a given context [10].

Initially, Python was used to clean the dataset. Following the standard procedure, it removed all non-English words and converted all text to lower case to ensure the program recognized all the exact words in the same manner. After this, we used Natural

Language Processing Tool Kit (NLTK) library [11] in Python to create a proper corpus using PorterStemmer and Stopword functions.

NLTK in Python has a list of stop words (such as 'the,' 'a', 'an') stored in 16 different languages. The Stopword function identifies these words in the corpus. PorterStemmer function is used to stem words; for example, after stemming, a word like "looking" becomes "look." Stemming is done on all words except stop words (removing stop words from the corpus). This process creates a good corpus, saves space in the database, and improves processing time. This corpus is used for classification and prediction purposes.

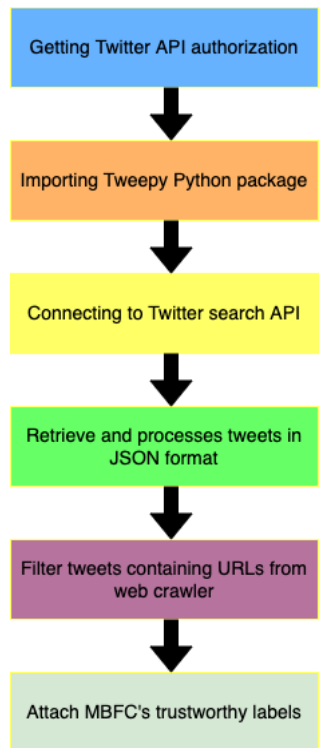

**Fig 1.** Process for collecting tweet data

We have used the Embedding Layer from Keras' library in Python [12] (which uses the word embedding technique for text-preprocessing – words with the same meaning will have similar representations). Before the data enters the Embedding Layer, it must be converted to a numerical form categorical, using One-Hot encoding [13]. After this, the timestamp is padded with zeros, either pre or post, to create sentences of equal length. Also, several feature vectors are defined for the Embedding Layer so the model can create vector representations based on the number. This is done so algorithms can generalize the corpus created to make any prediction.

We balanced the data because there were around 10 % more true tweets than fake tweets. Thus, we downsampled the true tweets to the same number as fake tweets. We removed URL patterns from the tweets, like https:// and other symbols and characters which resemble the part of a URL to have only normal text without any links to the tweet and then converted all the text to lowercase. The numbers of true and fake news are equally balanced into 289,826. Also, we divided the dataset randomly into 70% training and 30% testing. The total size of the training set is 405,756 (70%), and the testing set is 173,896 (30%).

### 3.4 Classifications

This section presents an overview of the models we used to classify our dataset.
We used the TensorFlow framework with a Python programming language to build the models. We also used other Python APIs like pandas, NumPy, matplotlib, and NLTK. TensorFlow has an embedding layer, GRU layer, LSTM layer, and dense, fully connected layer readily available along with Adam optimizer, which we used to build the models. After data preprocessing, the corpus created is classified by ensemble CNN+GRU and BiLSTM models.

We used cross-validation to verify that the models are not overfitting. We used a stratified K-Fold cross-validation technique where we had ten folds or splits to balance the classes in the training and testing datasets and to validate the models; We observed a mean accuracy of 91.6% for the ensemble CNN+Bi-GRU and 90.8% for Bi-LSTM models. Also, one of the advantages of using the ensemble CNN+Bi-GRU is that it reduces the chances of overfitting because it is the aggregate of all the model's output. The following sections show the structure of each model.

#### 3.4.1 Ensemble CNN+Bi-GRU Model

Gated Recurrent Units (GRUs) is a gating mechanism in recurrent neural networks, like the Long Short-Term Memory with a forget gate but fewer parameters than LSTM because it lacks an output gate. Figure 2 shows the architecture of GRU, where there are only two gates, which can also be called an update gate and reset gate. From Figure 2, the first gate from the input x, where there is the first sigmoid function from the left, is the reset gate and the following sigmoid function, which is also connected to the hidden state h to the right, is the update gate. GRU has been found to perform well with tasks like signal modeling and NLPs and works well with frequent datasets. So, a GRU can be generalized as a variation of LSTM because both have a similar design. GRUs use the update gate and the reset gate, the two primary operations in the GRU model. Therefore, these gates control the flow of information, which means that useful information can be kept, and unimportant information can be removed.

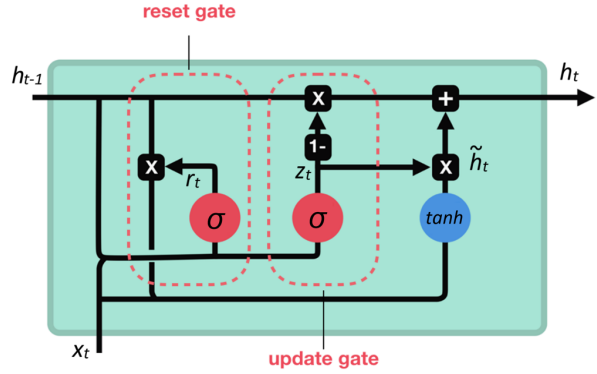

**Fig 2.** Gated Recurrent Unit (GRU)

Within deep learning, Convolutional Neural Network (CNN) is used to analyze structured data arrays (such as images) and is mainly used for image and text classification. Figure 3 shows the architecture of a CNN by stacking layers on top of each other in a sequence. These layers are usually convolutional, followed by activation and sometimes pooling layers. Additionally, CNNs are typically made up of 20 to 30 layers, with each layer capable of recognizing something more complex than the last. For example, using 3 to 5 layers, handwritten digits can be recognized, and with 25 layers, a human face can be recognized.

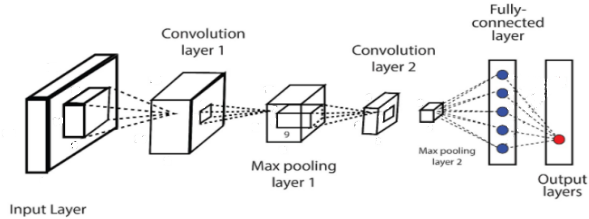

**Fig 3.** Convolutional Neural Network (CNN) model

After analyzing how efficient they can be for text classification and NLPs, we decided to use ensemble CNN+Bi-GRU, where we combine 1D CNN with a single Bi-GRU. 1D CNN performs better on text classification [14]. The Bi-GRU model works well on time-series data by looking at the earlier and later information sequences.

Figure 4 shows the layers of the ensemble CNN+Bi-GRU model we used for building CovidMis20 benchmark. There are two inputs for the model training purpose, one for CNN and one for Bi-GRU, followed by the embedding layer, convolutional layer, max pooling, dropout layer, and a dense or a fully connected layer with an activation function (sigmoid function). For GRU, the input layer is followed by the embedding layer, Bi-Directional Gru Layer, dropout layer, and a dense layer with a sigmoid function as an activation function. Then, we merged both the dense layers of CNN and Bi-GRU to form a dense layer (fully connected layer) with a sigmoid function.

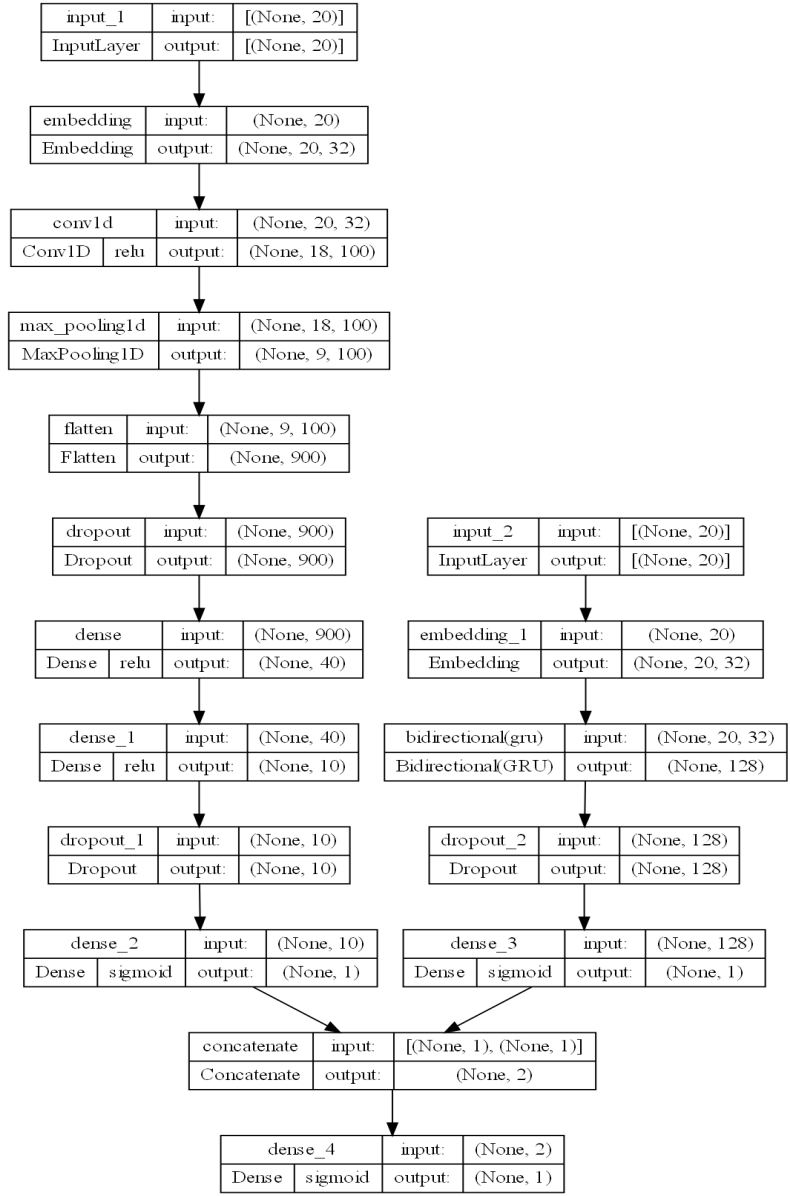

**Fig 4.** An ensemble CNN+Bi-GRU model

### 3.4.2 Bi-LSTM Model

Bi-LSTM is a type of RNN (Recurrent Neural Network [15]) used to avoid vanishing gradient problems. Bi-LSTM models have a cell state representing context information (storing information about past inputs for a specific time). Bidirectional LSTM can look at the data in both directions (left to right and right to left), so it is considered to store contextual information better than LSTM. Bidirectional LSTM is used when prediction depends on previous and future inputs. For example, a sentence like "I say Sam likes eating ___." Eating can be anything, so prediction depends on understanding the future and previous words. The Bi-LSTM model uses the time stamp technique to enter data into the model. The model learns the contextual information for the time stamp entered and tries to predict the next word in the sequence.

Bi-LSTM models and NLP (Natural Language Processing) use distributed representation techniques to describe the same data features across multiple scalable and interdependent layers. It tries to interpret or learn features of the same dataset on different scales (semantic similarities). Another reason to use the Bi-LSTM model is that around 2007, the models started to revolutionize speech recognition outperforming traditional models. Figure 5 shows the structure of an unfolded Bi-LSTM layer that contain the forward and backward LSTM layers.

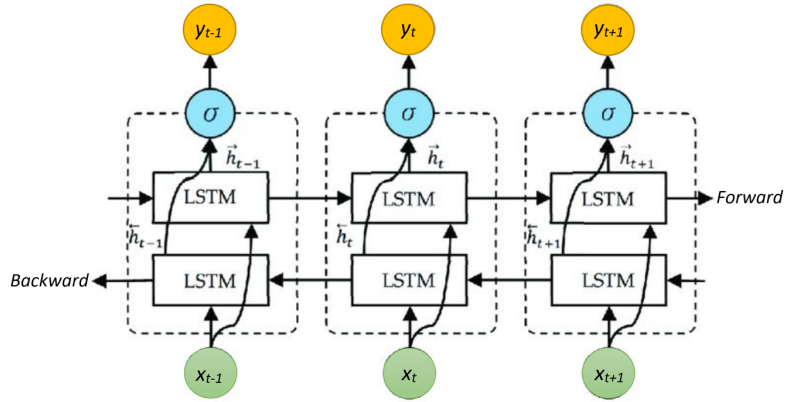

**Fig 5.** Unfolded architecture of BiLSTM with three consecutive steps

Figure 6 shows the layers of the Bi-LSTM model we used for this work. There is an input layer, followed by the embedding layer, Bi-Directional LSTM Layer, dropout layer, global max-pooling layer, and a dense layer with a sigmoid function as an activation function.

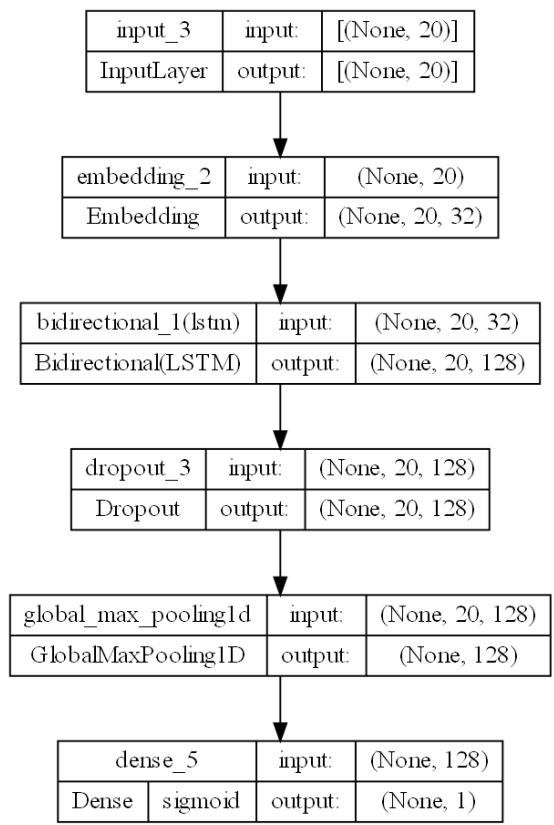

**Fig 6.** Bi-LSTM model

## 4 Results and Discussion

In this section, we provide details on the outcome of running the models on the Covid-Mis20 dataset.

Figure 7 shows the overall measurement results of the ensemble CNN+Bi-GRU and Bi-LSTM models to detect fake news from the CovidMis20 dataset. We have a 92.23% testing accuracy, 0.9025 precision (specificity), 0.945 recall (sensitivity), and 0.9232 F1-score with the ensemble CNN+Bi-GRU and 91.56% testing accuracy, 0.911 precision (specificity), 0.92 recall (sensitivity), and 0.9154 F1-score with the Bi-LSTM model. Figures 8 show the accuracy and validation accuracy for the ensemble CNN+Bi-GRU and Bi-LSTM, where they trained with a batch size of ten.

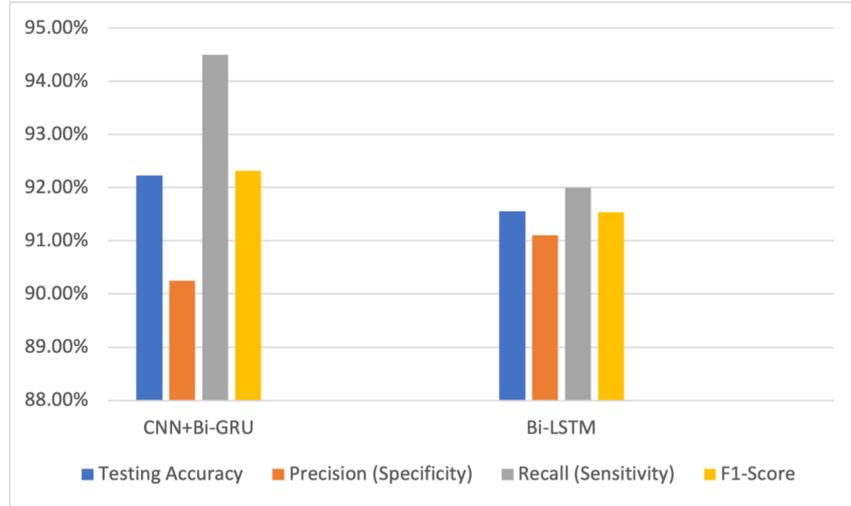

**Fig 7.** An ensemble CNN+Bi-GRU and Bi-LSTM performance

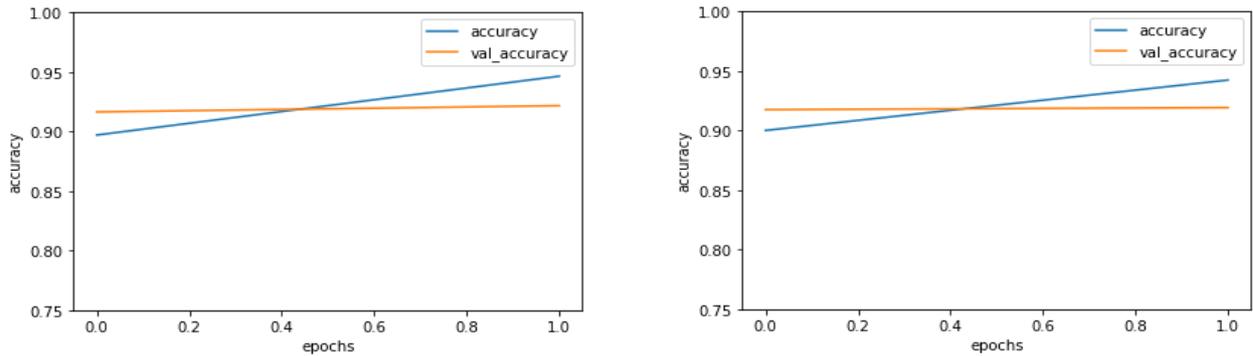

**Fig 8.** Evaluation phase performance of ensemble CNN+Bi-GRU (left) and Bi-LSTM (right)

Figure 9 shows the confusion matrix for ensemble CNN+Bi-GRU and BiLSTM models for text classification. We analyzed the confusion matrix in the evaluation phase, where 1 is fake and 0 is true. Thus, we can see a high accuracy of True Positive (TP) and True Negative (TN) percentages while detecting fake news from the Twitter dataset. For clarity, the meaning of true positive, true negative, etc. in our context is as follows:

- True Positive (TP): when predicted, fake news pieces are annotated as fake news.

- True Negative (TN): when predicted, true news pieces are annotated as true news.

- False Negative (FN): when predicted, true news pieces are annotated as fake news.

- False Positive (FP): when predicted, fake news pieces are annotated as true news.

Based on Figure 9, the ensemble CNN+Bi-GRU model classified 89.8% of fake tweets correctly and 10.2% incorrectly. At the same time, 5.5% of real tweets text was classified incorrectly by the model as fake, while classifying 94.5% of real tweets text. In contrast, the BiLSTM model correctly identified 91.1% of fake tweets text, misidentifying 8.9% as real. At the same time, 8.0% of real tweets' text was classified correctly and misclassified 92.0% incorrectly.

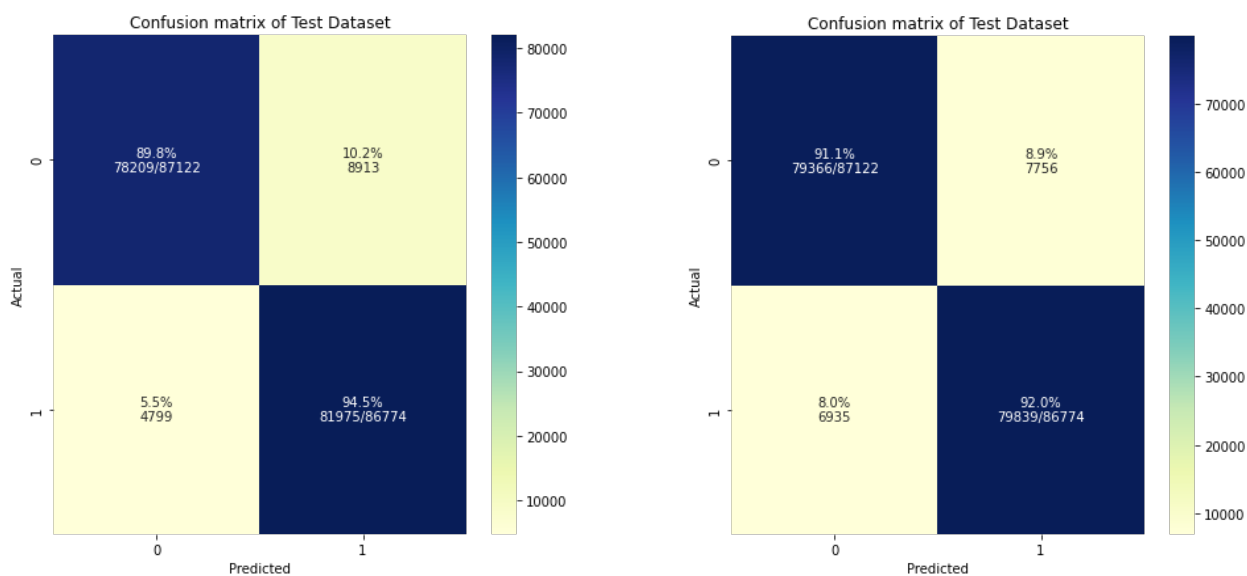

**Fig 9.** Confusion matrix of test dataset of ensemble CNN+Bi-GRU (left) and Bi-LSTM (right)

## 5      Conclusions and Future Work

This paper presented a comprehensive COVID-19 misinformation dataset, CovidMis20, which contains around 1,375,592 tweets from February to July 2020; CovidMis20 is class-wise balanced and can be used to develop automatic fake news detection models. Also, CovidMis20 is benchmarked by using deep learning models and projects them as potential baselines. The ensemble CNN+Bi-GRU model performs the best with 92.23% testing accuracy.

Although we used Covid-related Twitter tweets to build models to detect fake news, our techniques can easily be extended to other datasets for fake news detection.

Future work could be targeted toward using an enhanced evolutionary detection approach such as Particle Swarm Optimization (PSO), which aims to reduce the number of symmetrical features and obtain more accurate models.

## Acknowledgments

This work was supported in part by Saginaw Valley State University and the National Science Foundation under Grant OAC-2017289, National Institute of Health under Grant 1R15GM120820-01A1, and WMU FRACAA 2012-22.